\pdfoutput=1

\documentclass[11pt]{article}

\usepackage[final]{acl}

\usepackage{times}
\usepackage{tcolorbox}
\tcbuselibrary{skins, breakable, theorems}
\usepackage{latexsym}
\usepackage{multirow}
\usepackage{booktabs}
\usepackage{makecell}
\usepackage{cleveref}
\usepackage[T1]{fontenc}

\usepackage[utf8]{inputenc}

\usepackage{microtype}

\usepackage{inconsolata}

\usepackage{graphicx}
\usepackage{subcaption}
\usepackage{CJK}
\usepackage{algorithm} 
\usepackage{algpseudocode}
\usepackage{amsmath}

\newcommand{\tabincell}[2]{\begin{tabular}{@{}#1@{}}#2\end{tabular}}
\title{An Efficient and Precise Training Data Construction Framework for Process-supervised Reward Model in Mathematical Reasoning}

 \author{Wei Sun\textsuperscript{1,2}, Qianlong Du\textsuperscript{1}, Fuwei Cui\textsuperscript{1} \and Jiajun Zhang\textsuperscript{\dag}\textsuperscript{1,2,3}\\
         \textsuperscript{1}Institute of Automation, Chinese Academy of Sciences \\
         \textsuperscript{2}School of Artificial Intelligence, University of Chinese Academy of Sciences\\
         \textsuperscript{3}Wuhan AI Research\\
         \texttt{\{sunwei2023,fuwei.cui\}@ia.ac.cn}
         \texttt{\{qianlong.du,jjzhang\}@nlpr.ia.ac.cn} \\}

\begin{document}
\maketitle
\begingroup
\renewcommand\thefootnote{\dag}
\footnotetext{Corresponding author.}
\begin{abstract}
Enhancing the mathematical reasoning capabilities of Large Language Models (LLMs) is of great scientific and practical significance. Researchers typically employ process-supervised reward models (PRMs) to guide the reasoning process, effectively improving the models' reasoning abilities. However, existing methods for constructing process supervision training data, such as manual annotation and per-step Monte Carlo estimation, are often costly or suffer from poor quality. To address these challenges, this paper introduces a framework called \emph{EpicPRM} (\textbf{\underline{E}}fficient, \textbf{\underline{P}}rec\textbf{\underline{i}}se, \textbf{\underline{C}}heap), which annotates each intermediate reasoning step based on its quantified contribution and uses an adaptive binary search algorithm to enhance both annotation precision and efficiency. Using this approach, we efficiently construct a high-quality process supervision training dataset named \emph{Epic50k}, consisting of 50k annotated intermediate steps. Compared to other publicly available datasets, the PRM trained on \emph{Epic50k} demonstrates significantly superior performance. Getting Epic50k at \url{https://github.com/xiaolizh1/EpicPRM}.
\end{abstract}
\vspace{-4mm}


\begin{figure*}
    \centering
    \includegraphics[width=0.8\linewidth]{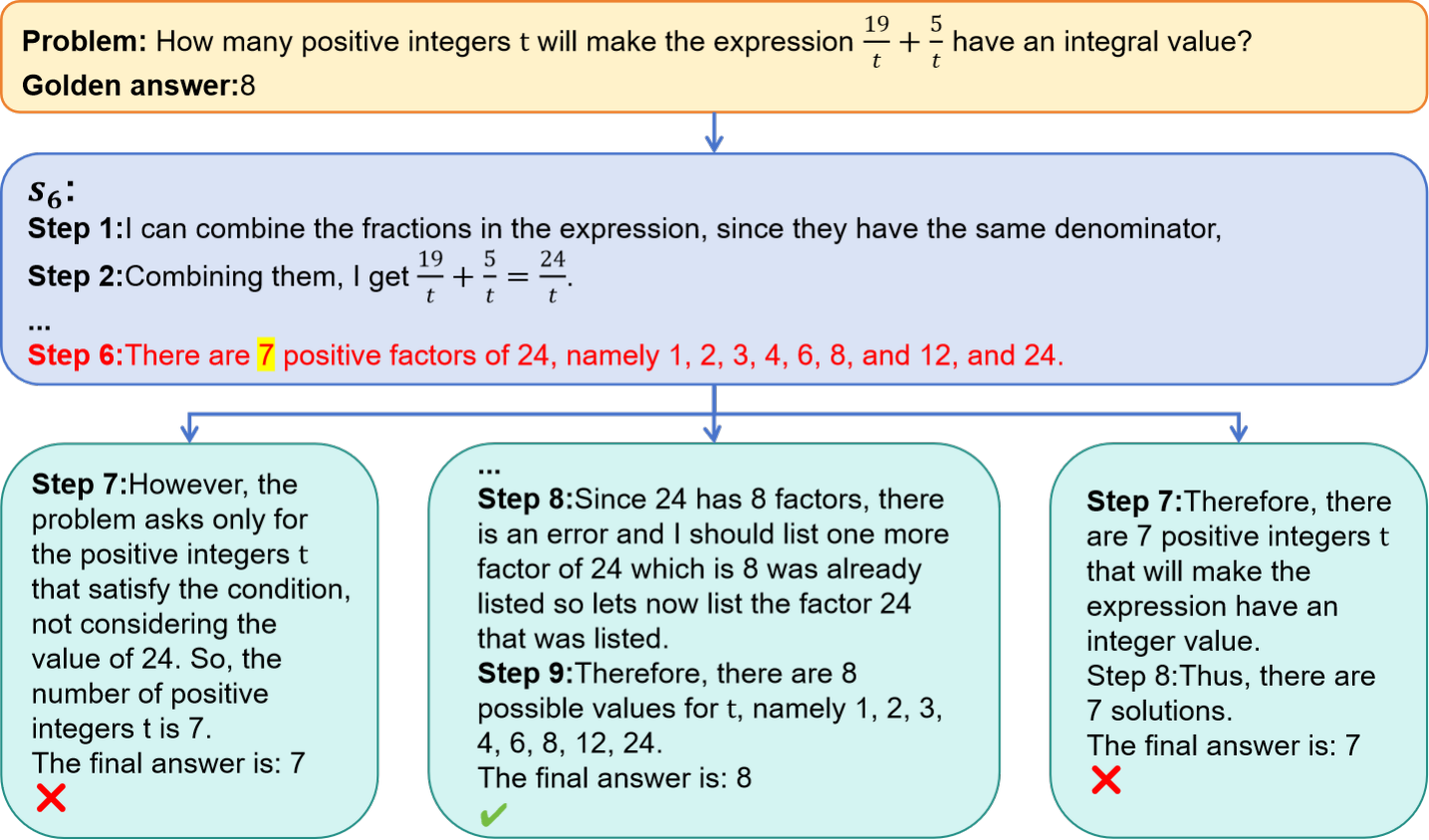}
    \caption{An examples of getting the correct answer based on the wrong steps. On the basis of \(s_{6}\), completer continues to sample a number of rollouts. We can see that since the completer has a certain error correction capability, even if there is an obvious error in step 6.}
    \label{fig:case}
    \vspace{-4mm}
\end{figure*}

\section{Introduction}

Improving the mathematical reasoning ability of Large Language Models (LLMs) can significantly accelerate advancements in artificial intelligence across various scientific domains. To enhance the mathematical reasoning capability of LLMs, several studies \cite{yu2023outcome, cobbe2021training, shao2024deepseekmath, chen2025lr} employ Outcome-supervised Reward Models (ORMs) to guide Chain-of-Thought (CoT) generation and leverage reinforcement learning to optimize generative performance. Among these approaches, process supervision, leveraging Process-supervised Reward Models (PRMs), has emerged as a more effective approach \cite{li2022making, uesato2022solving, lightman2023let, chen2024step}, providing substantial improvements through step-by-step supervision.

In process supervision, researchers typically construct an annotated dataset of reasoning processes and then train a PRM to guide the reasoning process. The quality of this annotated data is crucial for PRM effectiveness. Two primary approaches exist for dataset construction: manual annotation \cite{lightman2023let} and automatic annotation \cite{wang2024math, wang2024multi, luo2024improve}. While manual annotation ensures high data quality, it is labor-intensive and difficult to scale to diverse domains. Conversely, automatic annotation reduces labor costs but often results in lower annotation quality. This necessitates the use of larger datasets to compensate, which in turn demands significant computational resources.

To address these challenges, this paper introduces \emph{EpicPRM}, a fully open-source framework designed for the precise and efficient construction of high-quality annotated process supervision training datasets. Our approach leverages multiple LLMs with varying capabilities and from diverse sources to generate a more comprehensive set of CoT reasoning chains. We annotate intermediate steps by quantifying their contribution and identifying the first erroneous step within each chain. This identification is achieved through an adaptive binary search process, which adjusts the starting position and the number of Monte Carlo estimation samples based on problem difficulty. These optimizations result in a 64.39\% reduction in annotation costs compared to traditional sequential search algorithms. Using this framework, we constructed \emph{Epic50k}, a dataset containing 50k annotated intermediate steps. The PRM trained on \emph{Epic50k} achieves supervision performance comparable to, and in some cases surpassing, PRMs trained on the PRM800k and Math-Shepherd datasets, despite being less than 10\% the size.

Our main contributions are as follows:

1.  We propose a more efficient and precise framework for constructing process supervision data by quantifying the contribution of intermediate reasoning steps and optimizing the binary search algorithm.

2.  We publicly release \emph{Epic50k}, a high-quality process-supervised training dataset containing 50k intermediate reasoning steps.

3.  Our experimental results demonstrate the importance of data quality over quantity for training PRMs, showing that even a smaller, high-quality dataset can yield superior results.

\section{Background}
Existing methods for constructing process-supervision data first use LLMs to generate complete CoT chains, and then use Monte Carlo estimation to label the correctness of each step.

\textbf{Monte Carlo estimation.} Several related studies, including Math-Shepherd \cite{wang2024math}, MiPS \cite{wang2024multi}, and OmegaPRM \cite{luo2024improve}, have introduced a Monte Carlo (MC) estimation method to approximate the probability of obtaining a correct solution from a given state. This method designates a large language model (LLM) as a "completer," tasked with generating the subsequent reasoning steps from the current state \(s_{t}\) (\(s_{t} = [q; a_{1}; a_{2}; \dots; a_{t}]\), where \(q\) denotes the question and \(a_{1}\) through \(a_{t}\) represent the first \(t\) selected intermediate steps) until a final answer is reached. This generative process is termed a "rollout." After performing \(N\) independent rollouts, the number of successful rollouts \(M\) that arrive at the correct final answer is counted, and the probability of achieving a correct solution from the current state is estimated as \(M/N\).

\textbf{Identification of the first erroneous step.} In automated data construction, evaluating the correctness of intermediate steps within a complete erroneous CoT reasoning chain using MC estimation is computationally expensive. Prior work \cite{lightman2023let} has shown that effective PRM training only requires the identification of the first erroneous step in a complete CoT reasoning chain. Once the first erroneous step is identified, all preceding steps are annotated as correct, while subsequent steps are annotated as incorrect. To optimize this identification process, OmegaPRM \cite{luo2024improve} employs a binary search algorithm, reducing the number of MC estimations needed and thereby improving computational efficiency.
\section{Method}
In this section, we introduce how to make MC estimation more precise and how to find the first erroneous step more efficiently.
\subsection{Make MC Estimation More Precise}
\subsubsection{Perplexity Instead of Count}
Using MC estimation to represent the probability of obtaining the correct answer from a given state has inherent limitations. When \(N\) is not large enough, counting will be accidental, we argue that using \(M/N\) as a direct representation of the probability of sampling correct rollouts is imprecise. According to the Law of Large Numbers, estimating probabilities using counting methods necessitates a large number of samples to achieve reliable accuracy. For instance, observing two heads in two coin tosses does not imply a 100\% probability of obtaining heads. Similarly, if we sample $N$ rollouts and find $M$ correct ones, estimating the probability of correctness as $M/N$ inherently carries uncertainty due to sampling variability. By using perplexity, we can directly calculate the probability of the model generating each rollout, thereby mitigating errors introduced by sampling. Furthermore, if a correct rollout has an exceptionally low generation probability but is included in the sample, a counting-based method would overestimate the likelihood of obtaining correct rollouts. Therefore, we propose using perplexity (PPL) calculated by logarithmic probability to represent the sampling probability of each rollout:

\vspace{-4mm}
\begin{small} 
\begin{equation}
PPL\left (j;s_{t},\theta_{k} \right ) = \exp \left \{ -\frac{1}{L}\sum_{i=1}^{L}\log p_{\theta_{k}}\left ( x_{i} \mid x_{<i} \right ) \right \} \label{eq:ppl}
\end{equation}
\end{small}

where \(j\) represents a rollout, \(s_{t}\) represents the input prefix, \(\theta_{k}\) represents the \(k\)-th completer; \(p_{\theta_{k}}(x_{i} | x_{<i})\) is the probability that completer \(k\) generates the token \(x_{i}\) given the preceding tokens \(x_{<i}\); and \(L\) is the total number of tokens in rollout \(j\).

Consequently, each of \(K\) completers which independently samples \(N\) rollouts, with \(M\) of these rollouts yielding the correct answer, our MC estimate based on perplexity (PPL), denoted as \(MC_{PPL}\):

\vspace{-4mm}
\begin{small}
\begin{equation}
MC_{PPL}\left (s_{t},\theta_{1:K} \right ) = \frac{1}{K} \sum_{k=1}^{K} \frac{\sum_{m=1}^{M} \log_{}{PPL\left (j;s_{t},\theta_{k} \right )}}{\sum_{n=1}^{N} \log_{}{PPL\left (j;s_{t},\theta_{k} \right )} } \label{eq:mcppl}
\end{equation}
\end{small}

Compared with the counting, directly calculating the probability of obtaining correct answer eliminates the influence of accidental, that`s why we use \(MC_{PPL}\) instead of \(M/N\).

\subsubsection{Quantify Contribution of Steps}
\label{3.1.2}
\begin{figure}
    \centering
    \includegraphics[width=1\linewidth]{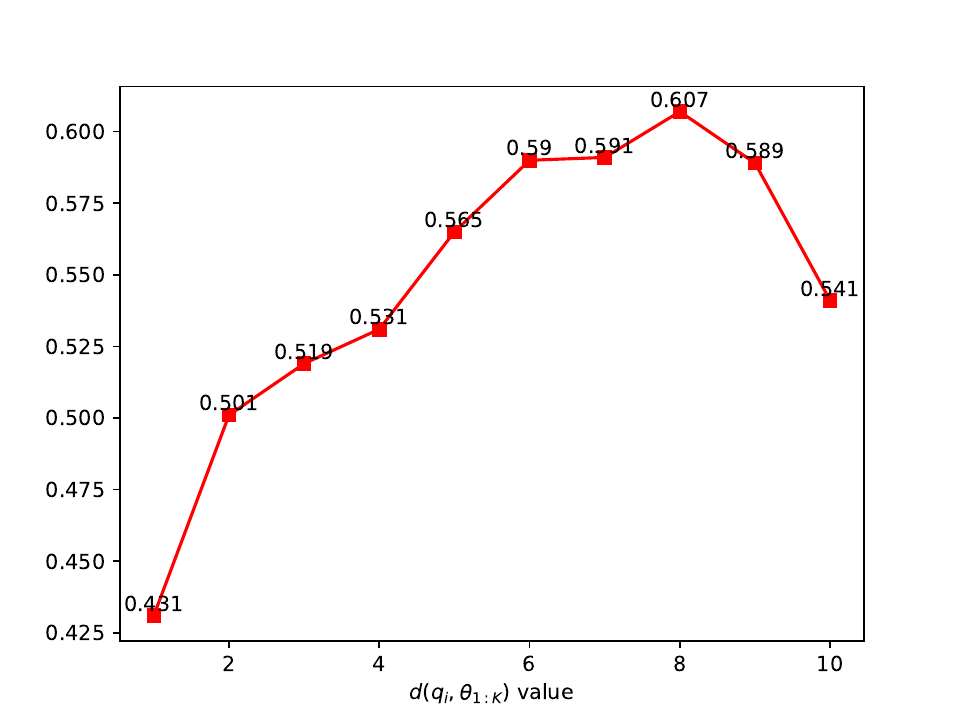}
    \caption{Relationship between the position of the first erroneous step and the difficulty of the problem. The horizontal axis represents the difficulty of the problem, and the vertical axis represents the ratio of the position of the first error step to the total number of steps.}
    \label{fig:diff}
    \vspace{-4mm}
\end{figure}
\textbf{Limitations of MC Estimation.} The prevailing automatic annotation methods operate under the assumption that if at least one of \(N\) sampled rollouts originating from state \(s_t\) produces the correct final answer (equivalent to \(MC_{PPL}\) value greater than 0), then all \(t\) intermediate steps within state \(s_t\) are considered correct. However, the inherent error correction capabilities of LLMs can obscure underlying inaccuracies in these intermediate steps. As demonstrated in Figure \ref{fig:case}, a completer may rectify an error at step \(a_t\) and still arrive at the correct final answer, despite \(a_t\) itself being flawed. This phenomenon results in false positives for \(a_t\), consequently diminishing the quality of the process supervision data and adversely affecting the supervisory efficacy of the trained PRM.

\textbf{Contribution-Based Annotation Method.} To annotate intermediate reasoning steps more precisely, we quantify the contribution of each step \(a_t\) to problem solving as the ratio of \(MC_{PPL}(a_t)\) to \(MC_{PPL}(a_0)\). This represents the ratio of the probability that the completer will continue to answer the question and ultimately arrive at the correct answer, given the first \(t\) steps (with the first \(t-1\) steps have been verified correct), to the probability that the completer will arrive at the correct answer without any reasoning steps. In Section \ref{section:3.1}, we specifically analyze the relationship between the quantified contribution \(C(a_{t};\theta_{1:K})\) and the correctness of the steps annotated by human expert through experiments and establish a \textbf{threshold \textbf{$\alpha$}} to be used as the criterion for annotating the steps:
\vspace{-2mm}
\begin{equation}
\begin{small}
\begin{aligned}
C(a_{t};\theta_{1:K}) & = \frac{MC_{PPL}\left (s_{t},\theta_{1:K} \right )}{MC_{PPL}\left (s_{0},\theta_{1:K} \right )} \\
Label(a_{t};\theta_{1:K}) & = \begin{cases}
0 & \text{if  }  C(a_{t};\theta_{1:K})  \le \alpha \\
1 & \text{if  } C(a_{t};\theta_{1:K}) > \alpha \\
\end{cases}
\end{aligned}
\end{small}
\end{equation}
\vspace{-3mm}
\vspace{-2mm}

\begin{figure}
    \centering
    \includegraphics[width=1\linewidth]{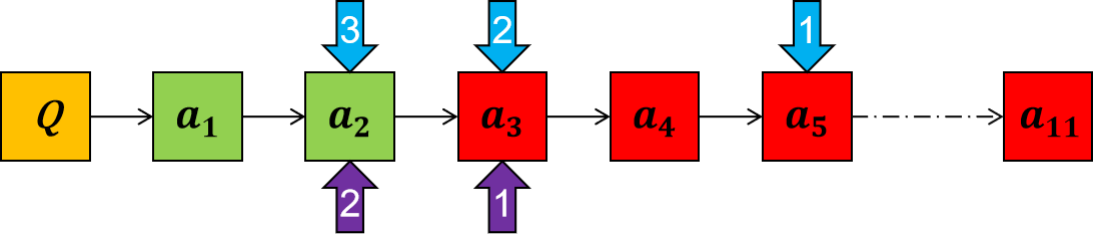}
    \caption{An example of comparing search processes between our adaptive binary search and traditional binary search. The blue arrow is traditional binary search, requiring 3 searches (starting from \(a_5\)) to find the first erroneous step \(a_3\). Conversely, our adaptive binary search (purple arrow) sets the initial search position to \(a_3\) based on question difficulty, requiring only 2 searches.
    }
    \label{fig:bs}
    \vspace{-4mm}
\end{figure}

\subsection{Adaptive Binary Search Algorithm}
In this subsection, we detail how to use the adaptive binary search algorithm to improve the efficiency of finding the first error step.
\subsubsection{Dynamic Starting Position} Human problem-solving behavior suggests that errors typically occur early in more complex problems, while simpler problems often involve calculation mistakes later in the solution. We hypothesize that LLMs follow similar patterns. To test this, we sampled \(22k\) erroneous rollouts and identified the position of the first erroneous step. We used the MC perplexity value of the initial problem state, \(MC_{PPL}(s_0)\), to represent problem difficulty—lower indicates greater difficulty. We classified problem difficulty \(d\left ( q_{i},\theta_{1:K}\right )\) into 11 levels:

\begin{small}
\begin{equation}
d\left ( q_{i},\theta_{1:K}\right ) = \text{round}\left( 10* MC_{PPL}\left ( s_{0},\theta_{1:K} \right ) \right) \label{eq:difficulty}
\end{equation}
\end{small}
\vspace{-4mm}

Our results, shown in Figure \ref{fig:diff}, reveal a strong correlation between the first erroneous step's position and problem difficulty. Based on these insights, we propose an enhanced binary search algorithm with a dynamic starting position, an example is shown in Figure \ref{fig:bs}. Instead of starting at \(T/2\), we adapt the starting point based on the problem’s difficulty \(d_{q_i}\) using Eq.(\ref{eq:mid}). This adjustment allows faster narrowing of the search range, reducing the number of MC estimations. The complete data construction process is shown in Algorithm \ref{algorithm1}.
\vspace{-4mm}

\begin{small}
\begin{equation}
mid = 
\begin{cases}
 \left\lfloor \frac{l+r}{2} \right\rfloor - \left\lfloor \frac{Len}{4} \right\rfloor & \text{if } d\left ( q_{i},\theta_{1:K}\right ) < 2 \\
 \left\lfloor \frac{l+r}{2} \right\rfloor & \text{if } 2 \leq d\left ( q_{i},\theta_{1:K}\right ) < 6 \\
 \left\lfloor \frac{l+r}{2} \right\rfloor + \left\lfloor \frac{Len}{4} \right\rfloor & \text{if } d\left ( q_{i},\theta_{1:K}\right ) \geq 6
\end{cases}
\label{eq:mid}
\end{equation}
\end{small}
\vspace{-2mm}

Where $d\left ( q_{i},\theta_{1:K}\right )$ represents the difficulty of \(q_{i}\) with K completers, \(r\) and \(l\) represent the upper and lower bounds of the binary search, and \(Len\) represents the total number of steps.

\vspace{-2mm}
\subsubsection{Dynamic Sampling Count} 

We observe that the computational overhead in automatic annotation is primarily attributable to the completer models' sampling of \(N\) rollouts. Existing methods employ a fixed value for \(N\) irrespective of problem difficulty, resulting in wasted computation on simpler problems and potentially insufficient sampling for complex problems, where the number of valid rollouts \(M\) at both \(s_0\) and \(s_t\) may be zero. To address this, we propose dynamically adjusting \(N\) based on problem difficulty. Specifically, we determine \(N\) at the initial state \(s_0\) (as defined in Eq.(\ref{eq:mcppl})) by initializing with a value of 16 and iteratively incrementing until \(M\) exceeds 10, thereby ensuring a minimum of 10 correct rollouts from each completer. This adaptive sampling strategy mitigates computational overhead while maintaining annotation quality.

\begin{small}
\begin{CJK}{UTF8}{gbsn} 
\begin{algorithm}[H]
    \floatname{algorithm}{Algorithm}
    \setcounter{algorithm}{0}
    \caption{Adaptive Binary Search for Annotating Process Supervision Data} 
    \hspace*{0.02in} {\bf Input:} Question $q_{i}$, response $A_{T} = (a_{1},...,a_{T})$, completers $\theta_{1:K}$, threshold coefficient $\alpha$ \\
    \hspace*{0.02in} {\bf Output:} The subscript of the first error in $A_{T}$
    \begin{algorithmic}[1]
        \State Initialize the lower bound $Left=0$, upper bound $Right=T-1$, flag for the first search $f=True$;
        \State Calculate the average $MC_{PPL}$ value and the threshold $H$ for error steps, $V=MC_{PPL}(s_{0},\theta_{1:k})$, $H=\alpha*V$;
        \While {$Left \le Right$} 
        \If {$f=True$ and $T \ge 4$}
            \State Compute $mid$;
        \Else
            \State $mid=\left \lfloor \frac{Left+Right}{2}\right \rfloor$;
        \EndIf
        \State $s_{mid}=(q_{i};a_{1},...a_{mid})$;
        \State $f=False$;
        \If {$MC_{PPL}(s_{mid},\theta_{1:K}) \le H$}
            \State $Right=mid-1$;
        \Else
            \State $Left=mid+1$;
        \EndIf
        \EndWhile
        \State return $Left$
    \end{algorithmic} 
\end{algorithm}
\label{algorithm1}
\end{CJK}
\end{small}

\vspace{-1mm}
\section{Experiments}
\vspace{-1mm}
\begin{figure}
    \centering
    \includegraphics[width=1\linewidth]{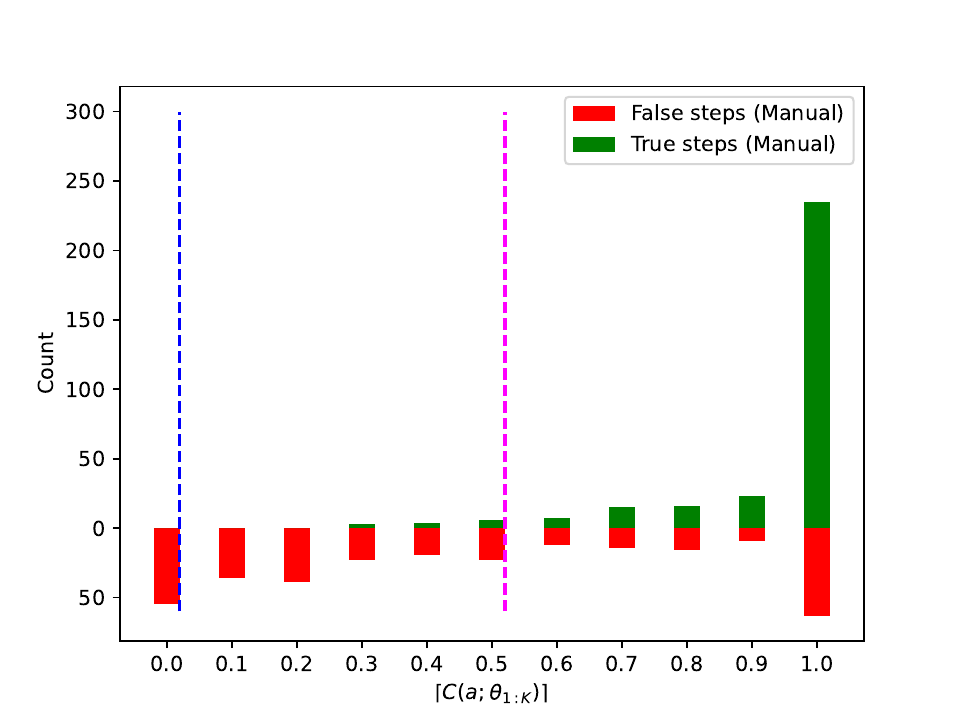}
    \caption{Relationship between manually annotated labels and step contribution. The x-axis depicts the quantified contribution per step, where the step values have been rounded up. All values greater than 1 are counted as 1. The vertical axis represents the number of manually annotated steps. The blue and pink dotted lines represent the existing method and our method respectively. The left side of dotted lines is annotated as the incorrect step, and the right side is annotated as the correct step. Our method yields more precise results in identifying correct and incorrect steps.}
    \label{fig:contribute}
    \vspace{-4mm}
\end{figure}

\subsection{Annotation Threshold Selection}
\label{section:3.1}

In this section, we experimentally determined the threshold in Section \ref{3.1.2}. We sampled 1,000 steps from the manually annotated PRM800k\cite{lightman2023let} and analyzed the ratio of their \(MC_{PPL}\) values at \(s_0\) and \(s_t\) to assess \(a_t\)'s contribution. Figure \ref{fig:contribute} shows that some manually annotated incorrect steps still had a positive ratio, indicating at least one correct rollout. This suggests that certain errors have minor negative effects, while some correct steps may introduce slight disruptions due to unconventional reasoning. As the ratio increases, correct steps become more prevalent while incorrect ones decrease. So our human experts have selected 0.5 as the threshold ($\alpha$) after random sampling validation.
As depicted in the bar chart of Figure \ref{fig:contribute}, indicate that a randomly chosen threshold within the range of 0 to 1 surpasses the discriminative capabilities of existing approaches, suggesting that human expert validation may not be necessary.

\subsection{PRM Training Data Construction}
\subsubsection{CoT Reasoning Chain Generation} We use the MATH dataset \cite{hendrycks2021measuring} to construct Epic50k. We adopted the same training and test set splits as PRM800k \cite{lightman2023let}, creating the MATH500 test set by selecting 500 representative questions from the original 5k-question MATH test set, and expanding the training set to 12k questions.
Initially, we randomly selected 3,500 questions from the training set, maintaining a level distribution ratio of 1:1:1:2:2 across levels 1 through 5. We then employed the LLaMA3-8B-Instruct \cite{touvron2023llama, llama3modelcard}, LLaMA3.1-8B-Instruct, and Qwen2-7B-Instruct \cite{yang2024qwen2} models to generate multiple complete solutions for these questions. LLaMA3-8B-Instruct was used for levels 1 and 2, while LLaMA3.1-8B-Instruct and Qwen2-7B-Instruct were used for levels 3 through 5. For each question, we selected 2 correct and 6 incorrect solutions exhibiting the lowest cosine similarity scores, and split each intermediate step of the solutions for automatic annotation.

\subsubsection{Intermediate Steps Annotation} When annotating intermediate steps, the \(MC_{PPL}\) values sampled by different LLMs can exhibit significant variability due to differences in their capabilities and biases. To mitigate this variability and more precisely reflect the true accuracy of state \(s_{t}\), we employed LLaMA3.1-8B-Instruct and Qwen2-7B-Instruct from multiple diverse sources as completers and calculated the average of their sampled \(MC_{PPL}\) values as the final estimate. As our model was not fine-tuned on mathematical datasets, we standardized the solution output format using a two-shot prompting approach (Appendix \ref{Prompt} for details). In total, we constructed a PRM training dataset containing 50,000 intermediate steps, with an approximate 1:1 ratio of correct to incorrect steps. To accelerate sampling, we utilized the vLLM inference framework \cite{kwon2023efficient}.The specific comparative analysis of step distribution can be found in the Appendix \ref{step_distribution}

\begin{figure*}[htbp]
\centering
 \begin{minipage}[t]{0.32\textwidth}
  \centering
  \includegraphics[scale = 0.33]{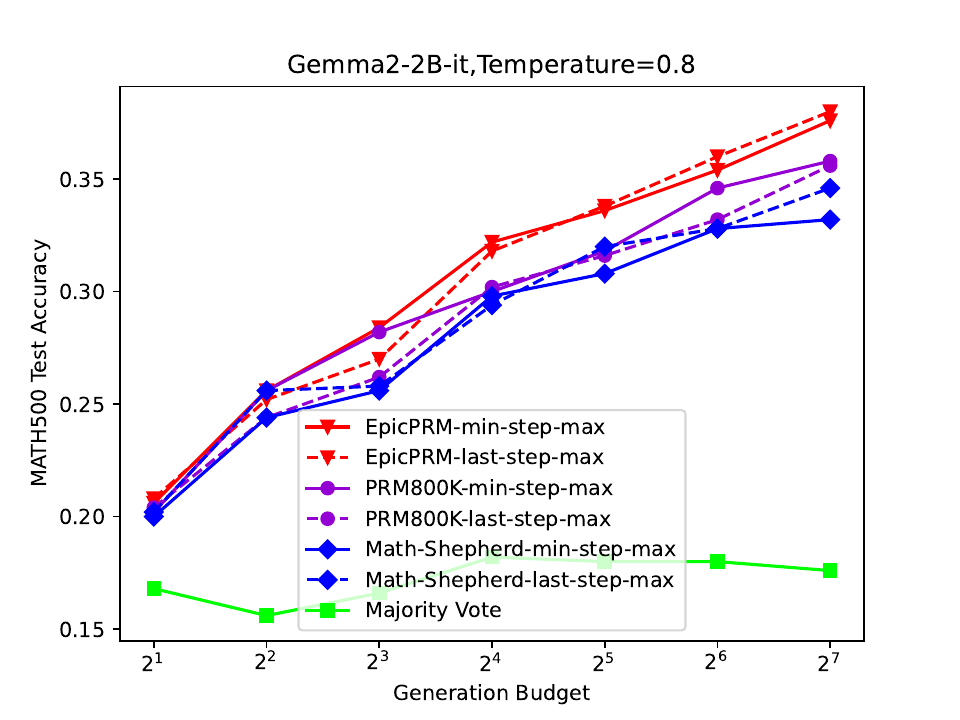}
  \subcaption{Gemma2-2B-it,Temperature=0.8}
  \label{gemma2b-0.8} 
 \end{minipage}
 \begin{minipage}[t]{0.32\textwidth}
  \centering
  \includegraphics[scale = 0.33]{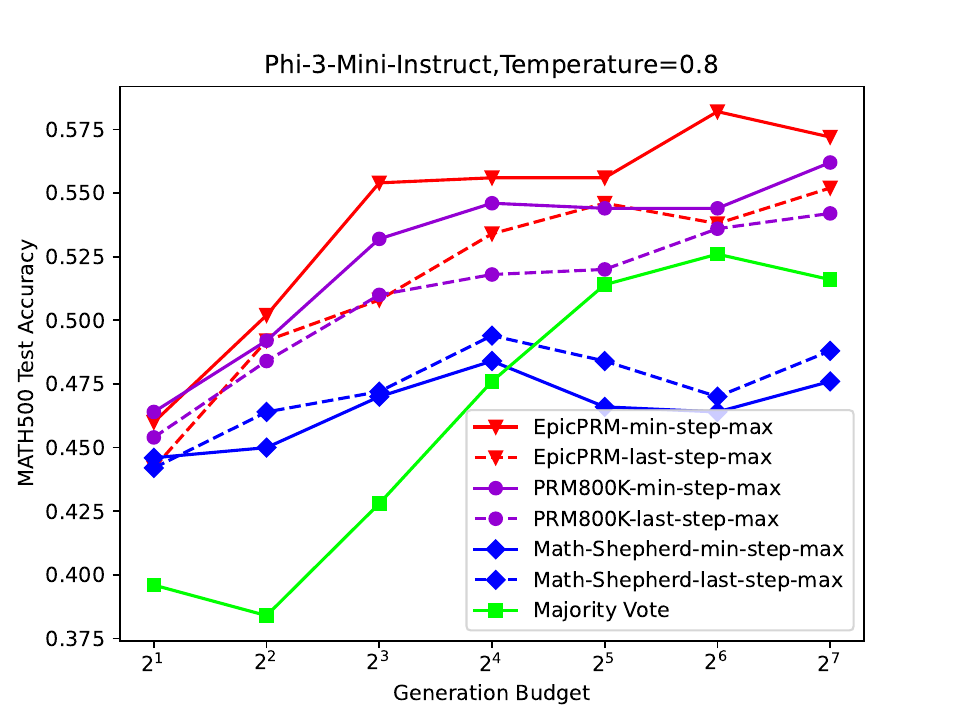}
  \subcaption{Phi-3-mini-4k-Ins,Temperature=0.8}
  \label{phi3-0.8} 
 \end{minipage}
 \begin{minipage}[t]{0.32\textwidth}
  \centering
  \includegraphics[scale = 0.33]{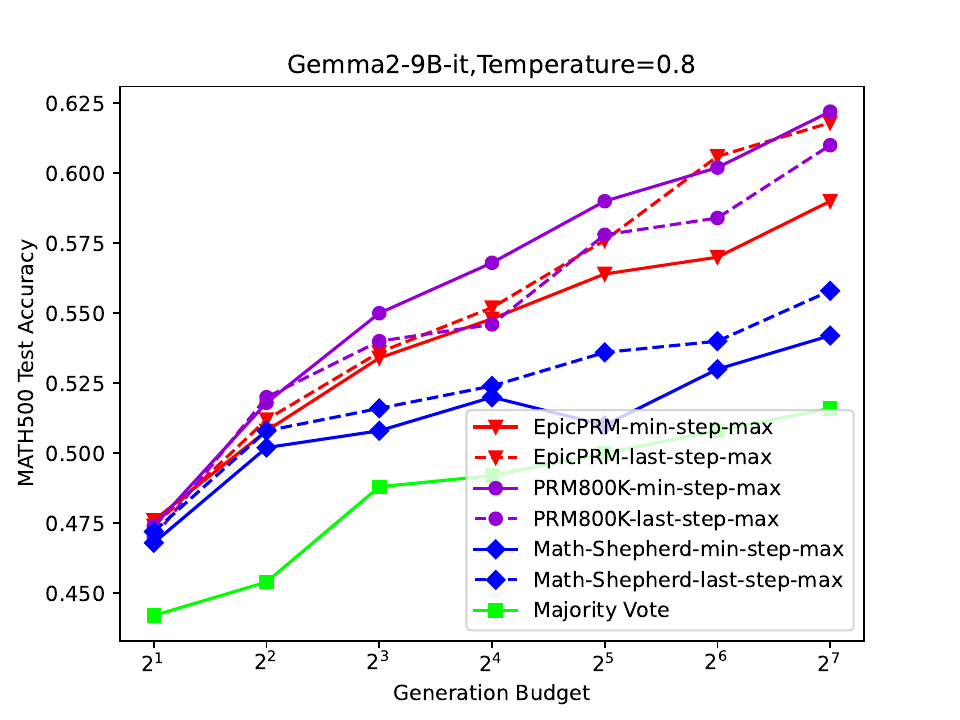}
  \subcaption{Gemma2-9B-it,Temperature=0.8}
  \label{gemma9b-0.8} 
 \end{minipage}
 \begin{minipage}[t]{0.32\textwidth}
  \centering
  \includegraphics[scale = 0.33]{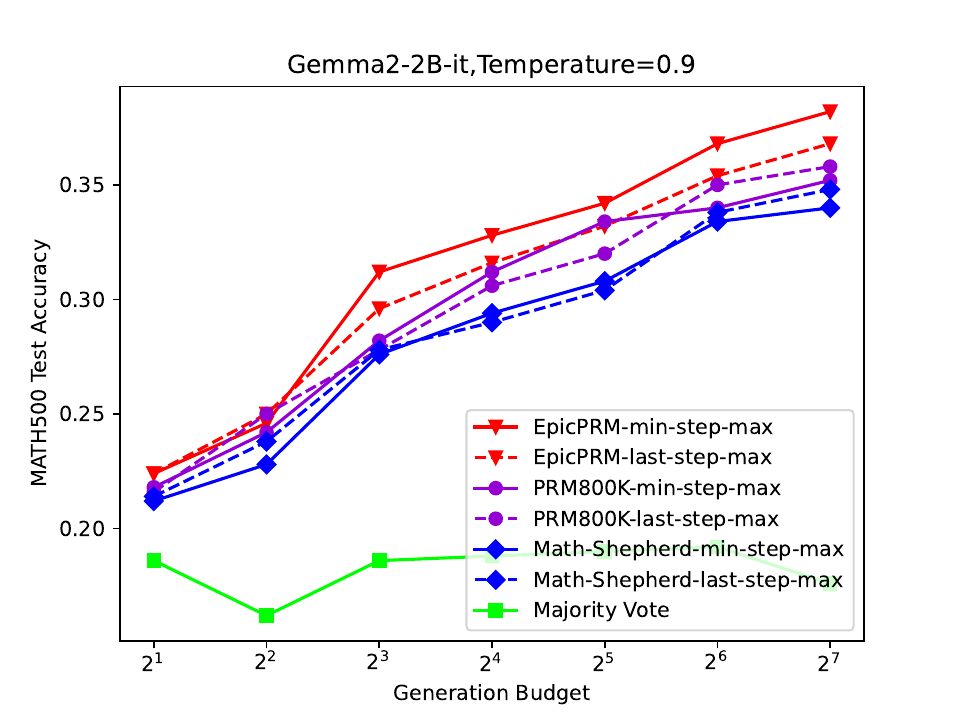}
  \subcaption{Gemma2-2B-it,Temperature=0.9}
  \label{gemma2b-0.9} 
 \end{minipage}
 \begin{minipage}[t]{0.32\textwidth}
  \centering
  \includegraphics[scale = 0.33]{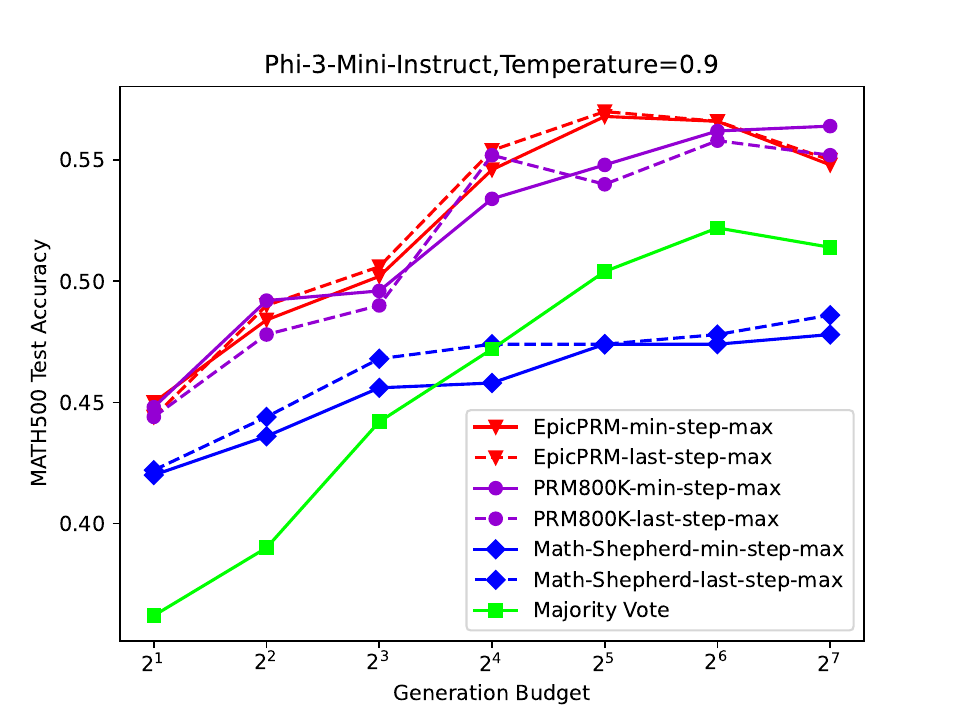}
  \subcaption{Phi-3-mini-4k-Ins,Temperature=0.9}
  \label{phi3-0.9} 
 \end{minipage}
 \begin{minipage}[t]{0.32\textwidth}
  \centering
  \includegraphics[scale = 0.33]{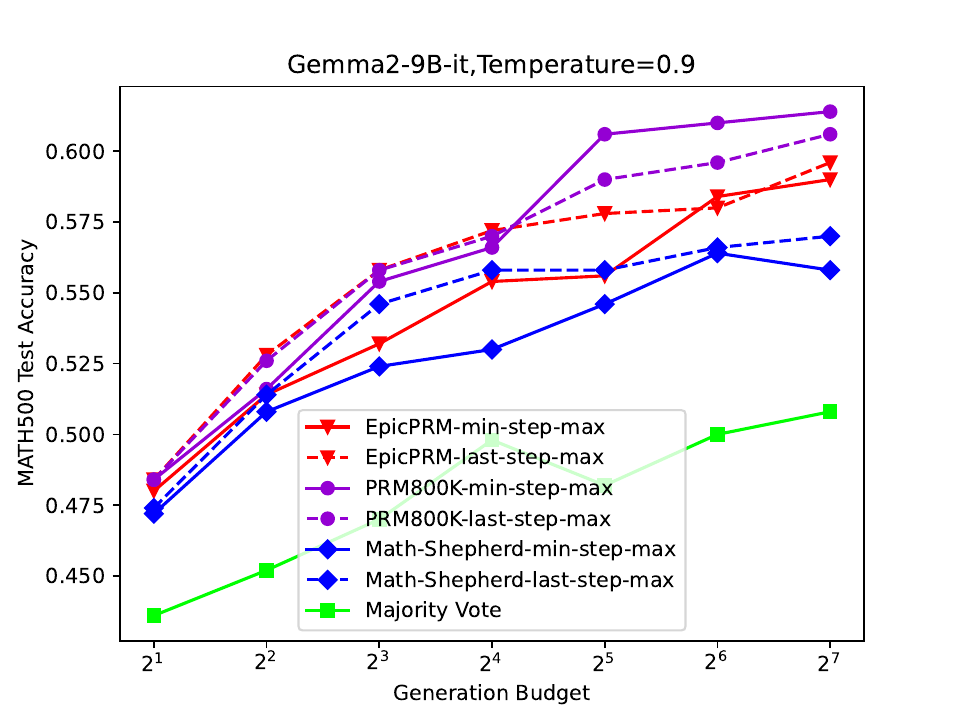}
  \subcaption{Gemma2-9B-it,Temperature=0.9}
  \label{gemma9b-0.9} 
 \end{minipage}
 \caption{Results of PRM weighted Best-of-N search for models of different scales at commonly used sampling temperatures.}
 \label{fig:bon_result}
 \vspace{-4mm}
\end{figure*}

\subsection{PRM Supervision Performance}
\subsubsection{PRM Training} Because our approach requires no additional supervised fine-tuning data, we chose Qwen2-math-1.5B-base \cite{yang2024qwen2}, a math-specialized LLM, as our base model. Unlike Math-Shepherd, which uses the complete solution sequence \(s_{1:T}\) as input and calculates softmax values over the vocabulary at special tokens of each step for loss computation, we treat PRM training as binary text classification which is same as PRM800k. We added a classification head to the base model's output layer and employ a standard binary classification loss:

\vspace{-3mm}
\begin{small}
\begin{equation}
Loss = \frac{1}{R} \sum_{i=1}^{N} \hat{y}_{i} \log y_{i} + \left( 1 - \hat{y}_{i} \right) \log \left( 1 - y_{i} \right)
\end{equation}
\label{eq:loss}
\vspace{-3mm}
\end{small}

Where \(\hat{y}_{i}\) represents the correctness label and \(y_{i}\) is the prediction score of the PRMs. \(R\) represents the number of training samples.

 When evaluating an intermediate step \(a_t\), we input only the preceding steps \(s_t\). This is preferable because inference-time iterative search algorithms (e.g., beam search) do not yet have subsequent content. We changed the format of the MATH-Shepherd dataset to a format suitable for binary text classification tasks.

\vspace{-2mm}
\begin{figure}
    \centering
    \includegraphics[width=1\linewidth]{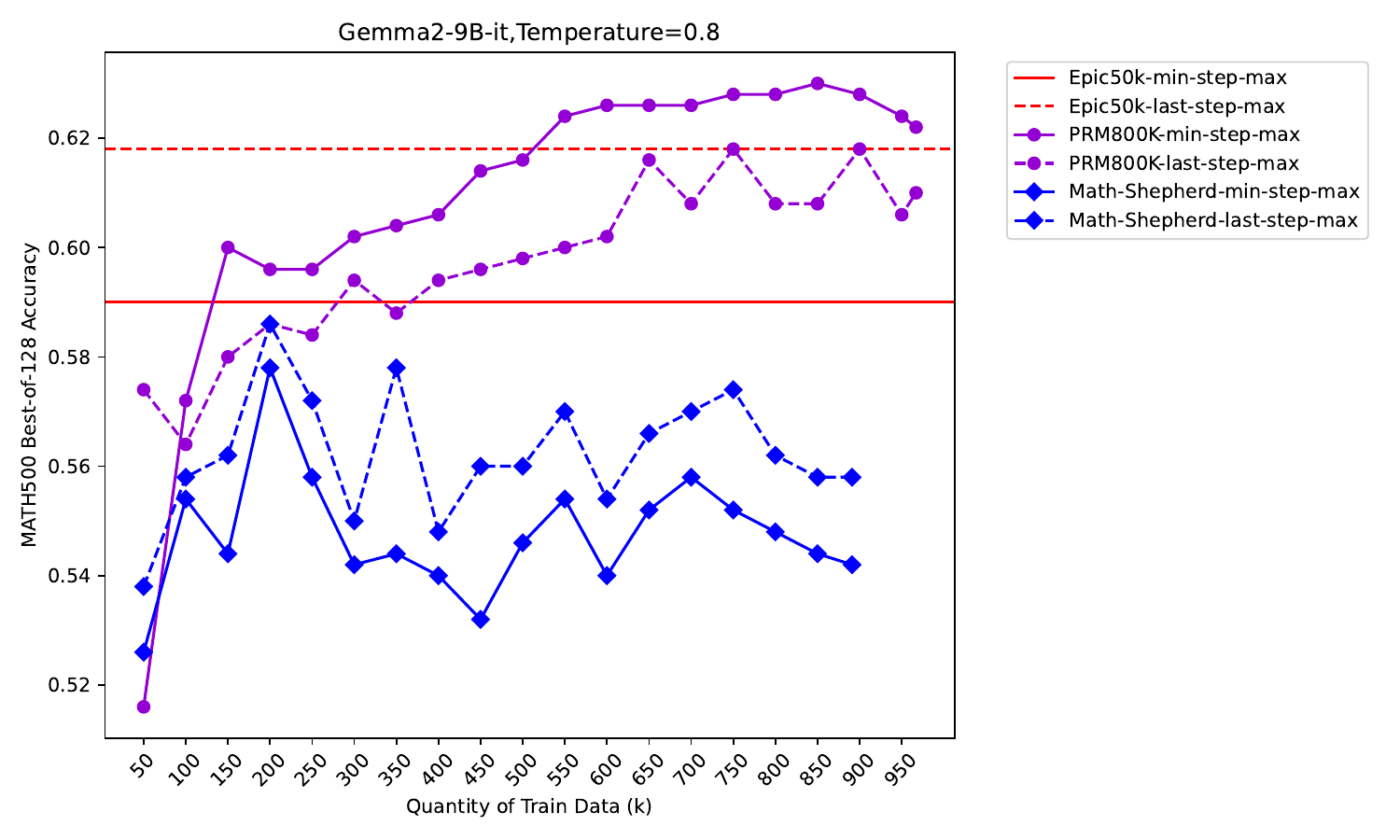}
    \caption{Comparison of the supervision performance of PRMs trained with varying datasets: subsets of PRM800k and MATH-Shepherd with at least 50k scales versus ours Epic50k.}
    \label{fig:sizes_accuracy}
\end{figure}
\subsubsection{Experiment Setup} 
We evaluated the effectiveness of the PRM on Gemma2-2B-it, Gemma2-9B-it\cite{team2024gemma}, Phi-3-mini-4k-Instruct (3.8B) \cite{abdin2024phi}, which are from different origins than the data generation models, using the \textbf{PRM weighted Best-of-N search} on MATH500. Specifically, a basic model was used as a generator, which sampled \(N\) complete solutions on the MATH500 test problem. We then used PRM to score these \(N\) solutions and selected one solution as the final answer. There are two selection strategies: one takes the minimum PRM score as the solution's score (\(Value=\min_{t\longrightarrow T} PRM(q_{i}|s_{t})\)); the other takes the PRM score of the last step as the solution's score (\(Value=PRM(q_{i}|s_{T})\)). Baseline process supervision datasets included PRM800k\cite{lightman2023let}, Math-Shepherd\cite{wang2024math}, and Self-Consistency\cite{wang2022self}. To align with our training methodology, we modified the Math-Shepherd data format accordingly. 

\vspace{-2mm}
\subsubsection{PRM Weighted BON Search Result}
Figure \ref{fig:bon_result} presents a comparison of PRM performance across various process-supervision datasets. Despite containing only 50k annotated intermediate steps, Epic50k demonstrates superior performance compared to Math-Shepherd across three distinct model sizes and sampling temperatures. Notably, Epic50k even surpasses PRM800k, which is trained on 800k manually annotated steps, in two of the evaluated models. Given the inherent precision of manually annotated data, it is unexpected that Epic50k, with a dataset size representing only 6.25\% of PRM800k, would consistently outperform it across all models. Table \ref{result_bo128} shows a comparative analysis of the Best-of-128 search results for different generators under varying PRMs supervisions. Epic50k`s consistent outperformance of existing automatic annotation methods sufficiently demonstrates the efficacy of our approach.

\begin{small}
\begin{table*}
\begin{center}
\begin{tabular}{cclcc}
\toprule[1pt]
\textbf{Generator Models} & \textbf{Size} & \textbf{Verifiers} & \textbf{Temperature} & \textbf{MATH500} \\
\hline
\multirow{5}{*}{\makecell{Gemma2-2B-it}} & \multirow{5}{*}{\makecell{2B}} & 2-shot CoT (Greedy) & 0 & 0.202 \\
& & +Self-Consistency mj@128 & \multirow{4}{*}{\makecell{0.8}} & 0.176 \\
& & +Math-Shepherd Best-of-128 &  & 0.346 \\
& & +PRM800k Best-of-128 &  & 0.358 \\
& & +Epic50k (Ours) Best-of-128 &  & \textbf{0.380} \\
\hline
\multirow{5}{*}{\makecell{Phi-3-mini-4k-Instruct}} & \multirow{5}{*}{\makecell{3.8B}} & 2-shot CoT (Greedy) & 0 & 0.378 \\
& & +Self-Consistency mj@128 & \multirow{4}{*}{\makecell{0.8}} & 0.516 \\
& & +Math-Shepherd Best-of-128 &  & 0.488 \\
& & +PRM800k Best-of-128 &  & 0.562 \\
& & +Epic50k (Ours) Best-of-128 &  & \textbf{0.572} \\
\hline
\multirow{5}{*}{\makecell{Gemma2-9B-it}} & \multirow{5}{*}{\makecell{9B}} & 2-shot CoT (Greedy) & 0 & 0.456 \\
& & +Self-Consistency mj@128 & \multirow{4}{*}{\makecell{0.8}} & 0.516 \\
& & +Math-Shepherd Best-of-128 &  & 0.558 \\
& & +PRM800k Best-of-128 &  & \textbf{0.622} \\
& & +Epic50k (Ours) Best-of-128 &  & 0.618 \\
\bottomrule[1pt]
\end{tabular}
\caption{\label{result_bo128} Comparison of the supervision effectiveness of PRMs trained on varying training datasets, evaluated using a Best-of-128 search with a temperature parameter set to 0.8.}
\end{center}
\vspace{-4mm}
\end{table*}
\end{small}

\begin{table*}[h]
  \centering
  \begin{tabular}{cccc}
    \toprule[1pt]
    \textbf{Algorithm} & \textbf{Verified steps} & \textbf{Sampled number}  & \textbf{Generated tokens}\\
    \hline
    Sequential Search  & 4204                      & 209.81K                     &76.67M              \\
    Binary Search      & 2620(-37.68$\%$)          & 133.78K(-36.24$\%$)         &42.52M(-44.54$\%$)  \\
    \textbf{Adaptive Binary Search(Ours)}      & \textbf{2539(-39.56$\%$)} & \textbf{70.40K(-66.45$\%$)} &\textbf{27.30M(-64.39$\%$)}  \\
    \bottomrule[1pt]
  \end{tabular}
  \caption{\label{Labeling costs}
    Cost of three search algorithms when annotating the same 800 erroneous steps.
  }
  \vspace{-4mm}
\end{table*}

\vspace{4mm}
\subsection{Comparison of Different Data Scales}

We compared Epic50k with randomly selected subsets of PRM800k and MATH-Shepherd using Gemma2-9b-it as the generator with a sampling temperature of 0.8, consistent with Figure \ref{gemma9b-0.8}. The results, shown in Figure \ref{fig:sizes_accuracy}, The horizontal axis represents the size of the randomly selected training data, while the vertical axis depicts the MATH500 Best-of-128 accuracy.
Epic50k consistently outperforms MATH-Shepherd across all data sizes and is only surpassed by PRM800k when the data exceeds 500k. This highlights Epic50k`s superior data efficiency. We posit that constructing a larger-scale training data using the EpicPRM framework would likely result in PRM supervision exceeding that of PRM800k. However, due to computational constraints, we limited our dataset to 50k.

\vspace{-2mm}
\subsection{Data Annotation Cost}
\vspace{-1mm}
\subsubsection{Compared Search Algorithms}
We also compared our method with the binary search algorithm and the sequential search algorithm with fixed sampling times to verify the advantage of our method in annotation efficiency. We annotated 800 randomly selected solutions. Both the binary search algorithm and the sequential search algorithm fixed the sampling times to 48, while our method sampled between 16 and 72 times until at least 10 correct solutions were sampled. Finally, we compared the number of sampling times and the number of tokens generated by the three search methods to annotate 800 incorrect solutions.

\begin{table*}[ht]
  \centering
  \vspace{1mm}
  \scalebox{0.95}{
    \begin{tabular}{lccccc}
    \toprule[1pt]
    \textbf{Model} & \textbf{GSM8K} & \textbf{MATH} & \tabincell{c}{\textbf{Olympiad-} \\ \textbf{Bench}} & \tabincell{c}{\textbf{Omni-} \\ \textbf{MATH}} & \textbf{Average} \\
    \hline
    Math-Shepherd-PRM-7B & 47.9  & 29.5  & 24.8  & 23.8  & 31.5  \\
    RLHFlow-PRM-Mistral-8B \cite{dong2024rlhf} & 50.4  & 33.4  & 13.8  & 15.8  & 28.4  \\
    RLHFlow-PRM-Deepseek-8B & 38.8  & 33.8  & 16.9  & 16.9  & 26.6  \\
    Skywork-PRM-1.5B \cite{liu2024skywork} & 59.0  & 48.0  & 19.3  & 19.2  & 36.4  \\
    Skywork-PRM-7B & 70.8 & 53.6  & 22.9  & 21.0  & 42.1  \\
    Qwen2-1.5B-PRM800k & 34.0 & 55.3  & 34.2  & 41.0 & 41.1  \\
    Qwen2-1.5B-MATH-Shepherd & 48.9 & 34.1  & 9.8  & 13.7  & 26.6  \\
    Qwen2-1.5B-Epic50k (Ours) & 55.6 & 36.1  & 20.2  & 30.0  & 35.5  \\
    Qwen2.5-1.5B-Instruct-PRM800k	& 49.1 &59.3	&32.5	&35.9	&44.2\\
    Qwen2.5-1.5B-Instruct-Shepherd	&44.1	&29.8	&12.1	&17.6	&25.9\\
    Qwen2.5-1.5B-Instruct-Epic50k	&58.1	&53.2	&32.2	&40.6	&46.0\\
    \bottomrule[1pt]
    \end{tabular}
    }
  \caption{\label{tab:processbench}
    Evaluation results of open-source process reward models (PRMs) on PROCESSBENCH. We report the F1 score of the respective accuracies on erroneous and correct samples.
  }
  \vspace{-4mm}
\end{table*}

\vspace{-2mm}
\subsubsection{Annotated Cost Comparison Results}
Table \ref{Labeling costs} presents the overhead compared with different search algorithms during annotating intermediate steps. Our adaptive binary search algorithm dynamically adjusts the number of samples and the initial search position based on the problem's difficulty. Compared to other search algorithms, it achieves significant reductions in 3 key metrics: the number of verified steps, total sampled number, and total generated tokens. In particular, the reduction in total generated tokens (a metric directly related to cost) demonstrates the strong effectiveness of our approach. We construct Epic50k using 4 NVIDIA A100 80G GPUs for approximately 260 hours, which is quite cost-effective and cheap.

In terms of validation steps, our method shows only a modest improvement over conventional binary search algorithms. This is because the average solution length in our dataset is just 10 steps, and most solutions can be completed with a standard binary search in fewer than 4 steps. Adjusting the initial search position provides noticeable advantages only for solutions with longer step lengths.

\vspace{-2mm}
\subsection{Out of Domain Generalization}

To evaluate our PRM on stronger generators and out of domain benchmarks, we conducted experiments on PROCESSBENCH \cite{zheng2024processbenchidentifyingprocesserrors}, a PRM assessment benchmark covering GSM8K, MATH, OlympiadBench, and Omni-MATH. Solutions were generated by LLaMA and Qwen models, including Qwen2.5-Math-7B/72B-Instruct, and labeled step-wise by human experts. Results are shown in Table \ref{tab:processbench}. Epic50k demonstrates strong generalization across benchmarks of varying difficulty. While it lags behind human-labeled PRM800k, it performs competitively with open-source PRMs, excelling at Olympiad-level problems where it matches Skywork-PRM-7B. Though Qwen2-1.5B-Epic50k scores slightly below Skywork-PRM-1.5B (by <1\%). Skywork leverages Qwen2.5-MATH-1.5B-Instruct, whereas our Qwen2-1.5B-Epic50k is based on the previous-generation Qwen2-MATH-1.5B-Base. Given the disparity in base models, this advantage is expected. Notably, despite being trained only on MATH, Qwen2-1.5B-Epic50k surpasses Skywork-PRM-1.5B on OlympiadBench and Omni-MATH, highlighting strong generalization from easier to harder problems. In order to be consistent with the base model of Skywork-PRM-1.5B, we conducted comparative experiments based on the Qwen2.5-1.5B-Instruct model. Our Epic50k performs significantly better than other methods.

\section{Related Work}
Improving the mathematical reasoning ability of Large Language Models (LLMs) can significantly accelerate advancements in artificial intelligence across various scientific domains, such as Translation~\cite{he2025r1}, Scientific agent reasoning~\cite{ren2025towards}, Long-text reasoning~\cite{wan2025qwenlong,chen2025ladm,chen2025lr,chen2025towards}, Multilingual~\cite{ghosh2025multilingual,yang2023bigtranslate}, Multi-modal~\cite{huang2025vision,jian2024large}.
\textbf{Prompt-based approaches.} Particularly Chain-of-Thought (CoT) \cite{wei2022chain}, have significantly advanced mathematical reasoning \cite{perez2021true}. Subsequent work has explored CoT fine-tuning \cite{ouyang2022training, li2024common, mitra2024orca, yu2023metamath} and extended prompt engineering to problem decomposition \cite{zhou2022least, hao2023reasoning} and programming \cite{chen2022program, li2023chain, zhou2023solving, chen2024alphamath, yang2024markov}. Selecting optimal solutions from multiple sampled paths has proven more effective than prompt engineering alone. Self-Consistency \cite{wang2022self} uses majority voting across generated CoT trajectories, while RAP \cite{hao2023reasoning} uses self-evaluation and feedback for sequential subproblem solving. However, existing research \cite{huang2023large} indicates that models encounter significant difficulty in self-assessing without external information.

\textbf{Application of PRMs.} Prior work \cite{huang2022large, wang2024math, cobbe2021training} has shown that using verifiers for problem-solving path assessment and selection is significantly more effective than self-consistency alone. PRMs have also been demonstrated to provide stronger supervision than ORMs \cite{lightman2023let}. While Math-Shepherd \cite{wang2024math} uses a specialized 
PRM, OVM \cite{yu2023outcome} employs iterative step sampling with a step-level verifier to identify optimal steps. Thus, high-quality process supervision data is crucial for training PRMs, which can further enhance generator models' mathematical abilities via methods like SFT \cite{zhang2024rest}, GROP \cite{shao2024deepseekmath}, and SVPO \cite{chen2024step}. Existing data construction methods include costly manual annotation (e.g., PRM800k \cite{lightman2023let}) and automatic annotation via Monte Carlo estimation (e.g., Math-Shepherd \cite{wang2024math}, MiPS \cite{wang2024multi}), with OmegaPRM \cite{luo2024improve} optimizing the latter using binary search within MCTS.

\section{Conclusion}
This study introduces EpicPRM, an open-source framework designed for the automatic annotation of intermediate reasoning steps in mathematical problem-solving. EpicPRM offers significant improvements in both precision and efficiency compared to existing automatic annotation methods. Notably, it achieves superior training results with less than 10\% of the data volume required by current state-of-the-art training datasets. 


DeepSeek-R1 \cite{deepseekai2025deepseekr1incentivizingreasoningcapability} attempted to apply process supervision but failed due to the high cost of PRM training data and its limited supervisory effectiveness. However, this does not disprove the feasibility of process supervision. In contrast, Microsoft`s rStar-MATH \cite{guan2025rstarmathsmallllmsmaster} demonstrated that models with 7B parameters or fewer can achieve reasoning capabilities comparable to OpenAI-o1 using process supervision. Our method further reduces PRM training costs while improving its supervisory effectiveness. This advancement provides more practical and efficient solutions for improving the reasoning capabilities of language models.

\section*{Limitations}

EpicPRM has made significant progress compared to existing automatic annotation methods, but it still has some limitations. Although we have significantly improved annotation precision, it has not yet reached the level of human expert annotation. Consequently, our automatically annotated data still contains some inherent noise. Furthermore, our current method relies on human-provided gold standard answers, meaning that complete independence from human supervision has not yet been achieved.

Theoretically, EpicPRM can be applied to any task that utilizes Chain-of-Thought (CoT) prompting and PRM-based process supervision. However, our experiment evaluation has been limited to the domain of mathematical reasoning. Therefore, the effectiveness and generalizability of EpicPRM in other reasoning domains, such as commonsense reasoning or logical inference, remain open questions. Future research will investigate the performance of EpicPRM across a wider range of tasks and domains to establish its broader applicability. This includes exploring potential adaptations or modifications that may be necessary to optimize performance in different contexts.

Given the high costs involved, we have not been able to perform a comprehensive analysis to assess the impact of adjusting the threshold in Section \ref{3.1.2} on data quality, nor have we explored the possibility of task-specific thresholds for optimal annotation. These inquiries will be the focus of our future work.

\section*{Acknowledgements}
We thank our colleagues Jianghao Chen, Jian Pu, Wen Yang, Cong Li, Tengxiao Xi, Tianyu Peng, Ziliang Pang, Junhong Wu for their insightful and constructive feedback. We thank Qian Li and Zhenggang Piao for their special assistance. In addition, we thank all reviewers for their valuable comments and recognition of our work. This work is supported by National Key R$\&$D Program of China N0.2022ZD0160602 and the Strategic Priority Research Program of Chinese Academy of Sciences under Grant N0.XDA04080400.

\bibliography{custom}

\newpage
\appendix
\section{Appendix}

\subsection{Prompt for 2-shot}
\label{Prompt}

\begin{tcolorbox}[colback=yellow!50!white,colframe=yellow!80!orange,title=\textcolor{black}{System prompt}]
\textcolor{red}{\textbf{<System>}} You are a powerful agent with broad math knowledge and good at accurate calculation on math equations.Below is an instruction that describes a task. Continue to finish the response that appropriately completes the request within a maximum of 40 steps. When outputting each step, mark the sequence number of each step at the beginning, and explicitly state the final answer after the final step following the format 'The final answer is:'. After outputting the final answer only once, be sure to stop outputting.
\end{tcolorbox}

\begin{tcolorbox}[colback=yellow!50!white,colframe=yellow!80!orange,title=\textcolor{black}{Example 1}]

\textcolor{red}{\textbf{<User>}} Instruction: If the lengths of two sides of a right triangle are 5 and 12 units, what is the least possible length, in units, of the third side? Express your answer in simplest radical form.

\textcolor{red}{\textbf{<Assistant>}} Response: Let's think step by step.

Step 1:I know that the Pythagorean theorem relates the lengths of the sides of a right triangle by the equation $a^2 + b^2 = c^2$, where c is the hypotenuse and a and b are the legs.

Step 2:Since I don't know which side is the hypotenuse, I'll try both possibilities and see which one gives me a smaller value for the third side.

Step 3:If I assume that the hypotenuse is $12$, then the other leg must satisfy $5^2 + b^2 = 12^2$, or $b^2 = 144 - 25 = 119$.

Step 4:Taking the square root of both sides, I get $b = \sqrt{119}$, which is already in simplest radical form.

Step 5:If I assume that the hypotenuse is the unknown side, then it must satisfy $5^2 + 12^2 = c^2$, or $c^2 = 25 + 144 = 169$.

Step 6:Taking the square root of both sides, I get $c = \sqrt{169} = 13$.

Step 7:Comparing the two values, I see that $\sqrt{119}$ is smaller than $13$, since $119$ is smaller than $169$.The final answer is $119$.

\end{tcolorbox}

\begin{tcolorbox}[colback=yellow!50!white,colframe=yellow!80!orange,title=\textcolor{black}{Example 2}]

\textcolor{red}{\textbf{<User>}} Instruction: A square has sides of length 10, and a circle centered at one of its vertices has radius 10. What is the area of the union of the regions enclosed by the square and the circle? Express your answer in terms of $\pi$.

Response: Let's think step by step.

Step 1:I want to find the area of the shaded region in this picture, where the blue is the square and the red is the circle.

Step 2:I notice that the circle and the square share a quarter of the circle's area, which is $\frac{1}{4}\pi r^2$, where $r = 10$.

\textcolor{red}{\textbf{<Assistant>}} Step 3:So I can subtract that from the sum of the areas of the circle and the square to get the area of the union.

Step 4:The area of the circle is $\pi r^2 = 100\pi$, and the area of the square is $s^2 = 100$, where $s = 10$.

Step 5:So the area of the union is $100\pi + 100 - \frac{1}{4}100\pi = 100 + \frac{3}{4}100\pi$.

Step 6:The final answer is: $100 + \frac{3}{4}100\pi$.
\end{tcolorbox}

\subsection{Step Distribution}
\label{step_distribution}
\begin{figure*}[h]
\centering
 \begin{minipage}[t]{0.3\textwidth}
  \centering
  \includegraphics[scale = 0.33]{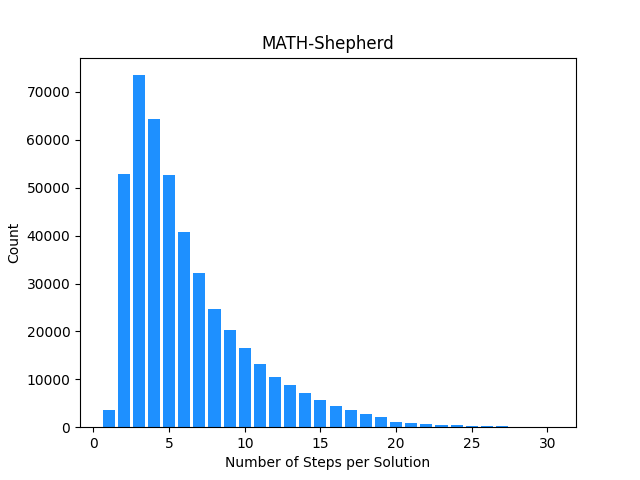}
 \end{minipage}
 \begin{minipage}[t]{0.3\textwidth}
  \centering
  \includegraphics[scale = 0.33]{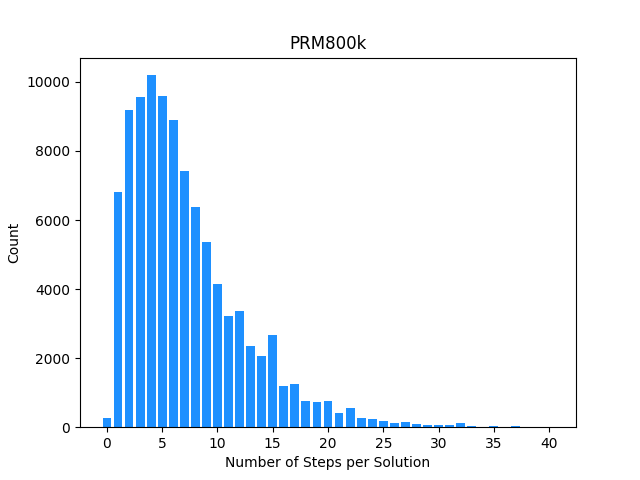}
 \end{minipage}
 \begin{minipage}[t]{0.3\textwidth}
  \centering
  \includegraphics[scale = 0.33]{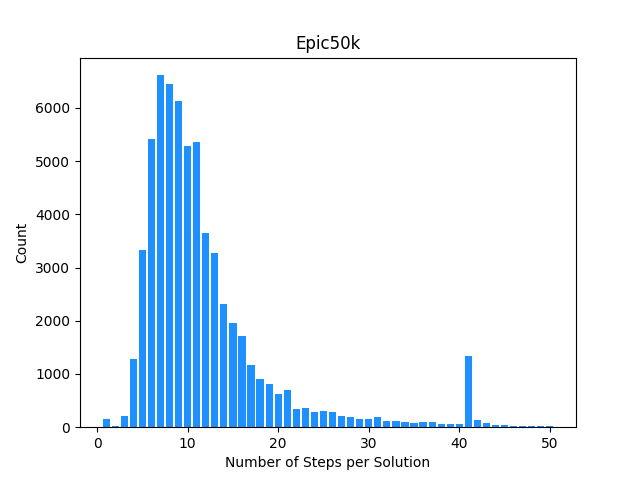}
 \end{minipage}
 \caption{Distribution of steps per solution.}
 \label{fig:total_step_analysis}
 \vspace{-4mm}
\end{figure*}

\begin{figure*}[t]
\centering
 \begin{minipage}[t]{0.3\textwidth}
  \centering
  \includegraphics[scale = 0.33]{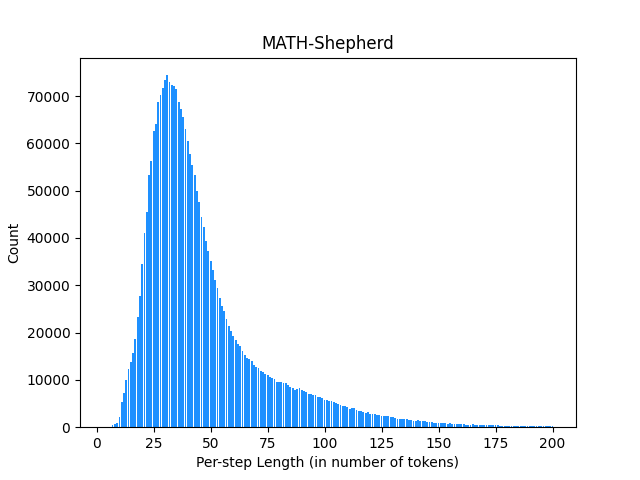}
 \end{minipage}
 \begin{minipage}[t]{0.3\textwidth}
  \centering
  \includegraphics[scale = 0.33]{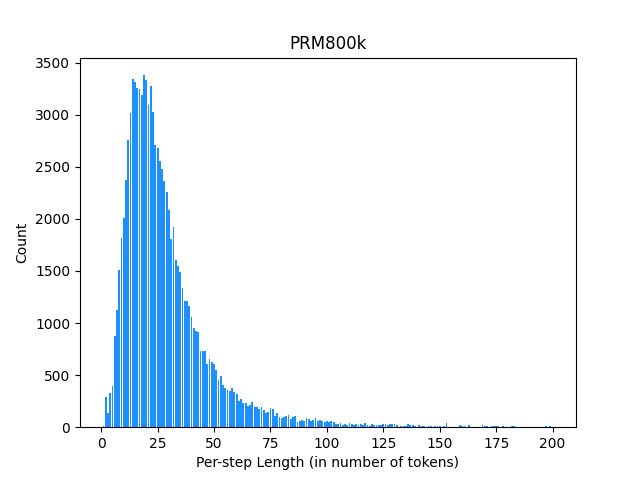}
 \end{minipage}
 \begin{minipage}[t]{0.3\textwidth}
  \centering
  \includegraphics[scale = 0.33]{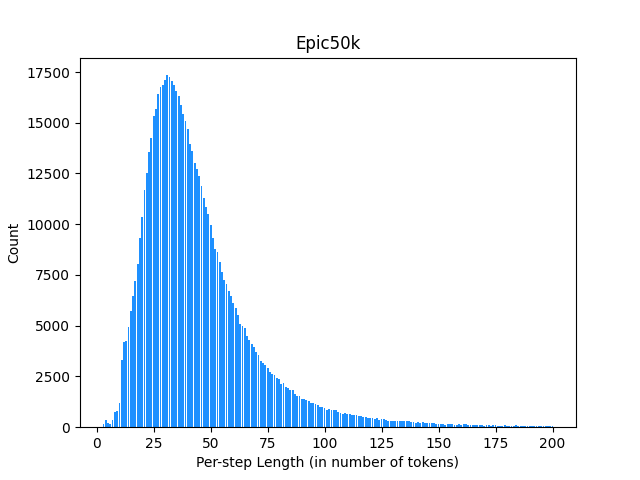}
 \end{minipage}
 \caption{Length distribution of each step.}
 \label{fig:step_len_analysis}
 \vspace{-10mm}
\end{figure*}

We analyzed the step distribution of three datasets-PRM800k, MATH-Shepherd, and our Epic50k-by examining both the number of steps per solution and the length of each step. Regarding the number of steps per solution (as illustrated in Figure \ref{fig:total_step_analysis}), our Epic50k dataset exhibits a concentration around 10 steps, whereas PRM800k and MATH-Shepherd are primarily clustered around 5 steps. This discrepancy is primarily attributed to the higher proportion of level 4-5 difficulty problems in Epic50k, which inherently necessitates more steps per solution. Subsequently, concerning the length of each step (shown in Figure \ref{fig:step_len_analysis}), we observe that Epic50k and MATH-Shepherd exhibit step lengths predominantly ranging from 25 to 50 tokens. In contrast, PRM800k demonstrates a comparatively shorter step length, concentrated around approximately 25 tokens.

\end{document}